\documentclass{article}

\PassOptionsToPackage{numbers, sort&compress}{natbib}

\usepackage[dblblindworkshop,final]{neurips_2025}

\usepackage[utf8]{inputenc} 
\usepackage[T1]{fontenc}    
\usepackage{hyperref}       
\usepackage{url}            
\usepackage{booktabs}       
\usepackage{amsfonts}       
\usepackage{nicefrac}       
\usepackage{microtype}      
\usepackage{xcolor}         
\usepackage{graphicx}
\usepackage{multirow}
\usepackage{enumitem}
\usepackage{tabularx}
\usepackage[capitalize]{cleveref}
\usepackage{pifont}
\usepackage{wrapfig}

\usepackage[detect-all,separate-uncertainty = true]{siunitx}
\makeatletter
\renewcommand{\paragraph}{%
  \@startsection{paragraph}{4}%
  {\z@}{1.5ex \@plus 1ex \@minus .2ex}{-0.5em}%
  {\normalfont\normalsize\bfseries}%
}
\makeatother 

\newcommand{\benchmarkname}{BioBench}
\newcommand{\nckpts}{$46$}
\newcommand{\nfamilies}{$11$}

\newcommand{\verticalcell}[1]{\multicolumn{1}{c}{\rotatebox{90}{#1}}}

\newcommand{\cmark}{\ding{51}}
\newcommand{\xmark}{\ding{55}}

\setlist{nosep}

\title{\benchmarkname{}: A Blueprint to Move Beyond\\ImageNet for Scientific ML Benchmarks}

\author{%
  Samuel Stevens \\
  The Ohio State University\\
  \texttt{stevens.994@osu.edu} \\
}

\begin{document}

\maketitle

\begin{abstract}
ImageNet-1K linear-probe transfer accuracy remains the default proxy for visual representation quality, yet it no longer predicts performance on scientific imagery.
Across \nckpts{} modern vision transformer checkpoints, ImageNet top-1 accuracy explains only \num{34}\% of variance on ecology tasks and mis-ranks \num{30}\% of models above \num{75}\% accuracy. 
We present \benchmarkname{}, an open ecology vision benchmark that captures what ImageNet misses. 
\benchmarkname{} unifies \num{9} publicly released, application-driven tasks, \num{4} taxonomic kingdoms, and \num{6} acquisition modalities (drone RGB, web video, micrographs, in-situ and specimen photos, camera-trap frames), totaling \num{3.1}M images. 
A single Python API downloads data, fits lightweight classifiers to frozen backbones, and reports class-balanced macro-F1 (plus domain metrics for FishNet and FungiCLEF); ViT-L models evaluate in 6 hours on an A6000 GPU. 
\benchmarkname{} provides new signal for computer vision in ecology and a template recipe for building reliable AI-for-science benchmarks in any domain. 
Code and predictions are available at \href{https://github.com/samuelstevens/biobench}{github.com/samuelstevens/biobench} and results at \href{https://samuelstevens.me/biobench}{samuelstevens.me/biobench}.
\end{abstract}

\section{Introduction}

Machine learning now drives everything from protein structure prediction to planetary-scale biodiversity surveys, yet progress depends on benchmarks that tell us which models to trust.
Vision research still orients around ImageNet-1K, MS COCO, and ADE20K \citep{deng2009imagenet1k,lin2014mscoco,zhou2017ade20k}, and state-of-the-art claims like vision transformers \citep{dosovitskiy2020vit}, self-supervised pre-training \citep{oquab2023dinov2} or image-text pre-training \citep{radford2021clip} are routinely justified by gains on those leaderboards.

Scientific images, however, are \textit{not} web photographs.
Radiographs and histopathology slides emphasize internal or cellular structure \citep{zech2018variable}; microbiology depends on high-magnification micrographs of microorganisms \citep{raghu2019transfusion}; and ecology relies on camera-trap or specimen imagery in uncontrolled environments \citep{tuia2022perspectives,weinstein2018computer}.
These sources differ in content, scale, and acquisition method from the datasets that govern general computer vision progress.

\begin{figure}[t]
    \centering
    \includegraphics[width=\linewidth]{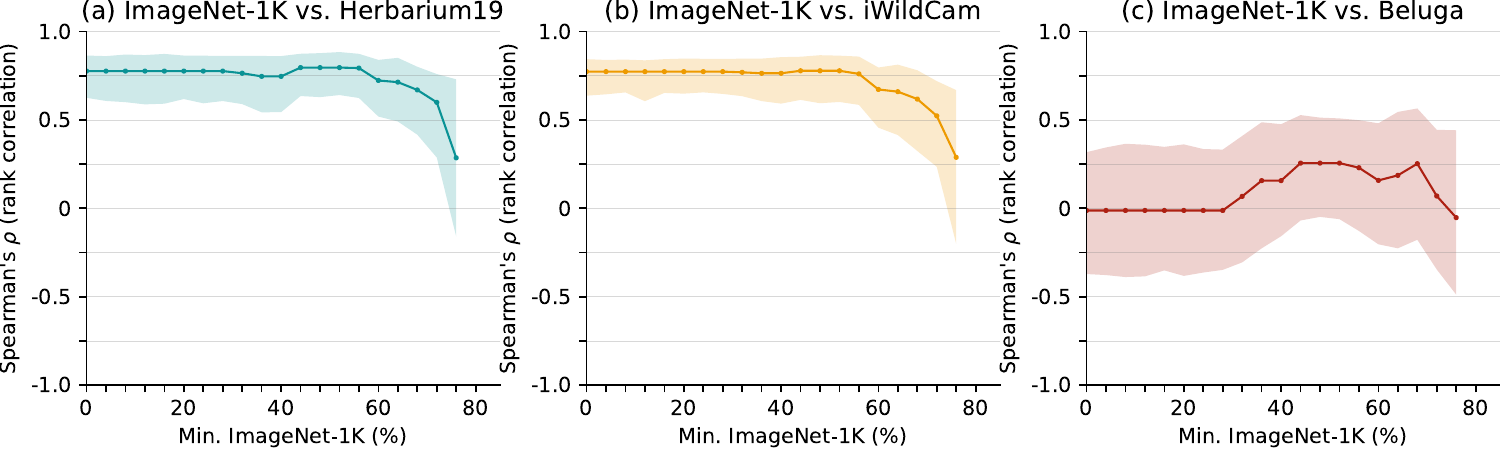}
    \caption{
    Predictive validity of ImageNet-1K accuracy across (a) species classification of plants \citep[Herbarium19,][]{tan2019herbarium19}, (b) species classification of animals in camera trap images \citep[iWildcam,][]{beery2020iwildcam,koh2021wilds} and (c) individual identification of beluga whales \citep[Beluga,][]{algasov2024beluga,vcermak2024wildlifedatasets} measured with Spearman's rank correlation coefficient $\rho$ between ImageNet-1K and task rankings, computed across all checkpoints with ImageNet Top-1 accuracy $\geq T\%$ (x-axis).
    Shaded region shows \num{95}\% bootstrapped confidence intervals. 
    \textbf{ImageNet-1K fails to predict model rankings on specific tasks as models improve.}
    }\label{fig:hook}
    \vspace{-12pt}
\end{figure}

The mismatch is not merely cosmetic. 
Across three publicly released ecology tasks (long-tail species ID \citep{garcin2021plantnet300k}, drone-video behaviour recognition \citep{kholiavchenko2024kabr}, and specimen trait inference \citep{khan2023fishnet}) we measure Spearman's rank correlation coefficient $\rho$ between ImageNet-1K top-1 accuracy and task accuracy for \nckpts{} modern computer vision checkpoints spanning supervised \citep{wightman2021resnet}, self-supervised \citep{oquab2023dinov2}, and image–text \citep{radford2021clip,zhai2023siglip,fini2024aimv2} pre-training objectives.
Once models surpass the now-common \num{75}\% ImageNet threshold, Spearman's rank correlation $\rho$ falls below \num{0.25} (see \cref{fig:hook}).
Generic benchmark accuracy, long used as a barometer of visual understanding, stops predicting performance on the scientific tasks we measure once models clear the 75\% ImageNet top-1 threshold.
Other work hints that the same ``ranking cliff'' afflicts other real-world tasks such as \citep{fang2023does,vishniakov2024beyondimagenet}.
Because ecological domains offer both scientific diversity and abundant open data, they provide an ideal testbed to systematically investigate how benchmark predictivity fails under realistic distributional shift.

ImageNet fails for two intertwined reasons. 
First, \textit{distribution mismatch}: its RGB web photos share neither spectrum nor noise with camera-trap infrared, multi-spectral drone passes, or microscope slides, so models optimized for ImageNet seldom work for scientific imagery.
Second, scientific tasks are fine-grained and long-tailed: ecologists distinguish thousands of insect species, pathologists dozens of rare tumor sub-types; ImageNet’s \num{1000} classes contain few such subtle distinctions and are heavily skewed toward frequent objects. 
Together these gaps explain why increasing ImageNet accuracy ceases to improve performance once models venture into application-driven tasks \citep{rolnick2024applicationdriven}.

The obvious remedy is to benchmark models on the applications themselves. 
When tasks are drawn directly from practice, their image distributions align by construction, their labels inherit the domain’s natural granularity, and their objectives mirror the questions scientists actually ask. 
Many fields still lack shared datasets of this sort, but ecology is an exception: years of CV4Ecology challenges have produced public tasks for species identification, behavior recognition, and trait inference. 
By consolidating these efforts into a single suite we can test whether application-driven benchmarks restore predictive power and provide a template for other scientific domains.

%
%
%
%

Why, then, has no unified benchmark appeared?
Because three hurdles discouraged even the most committed researchers. 
First, fragmentation: every ecology dataset shipped in its own repository with idiosyncratic file trees, splits, and metric scripts. 
Second, perceived sufficiency: most vision researchers assumed that strong ImageNet accuracy, averaged over scattered per-task leaderboards, already served as an adequate proxy, so consolidating tasks seemed low-yield. 
Finally, non-overlapping waves of progress: benchmarks surfaced one at a time; every release compared against the ``best'' backbone of that moment and the authors’ favorite tricks. 
Because checkpoints, hyper-parameters, and evaluation scripts kept changing, nobody could tell whether any single model genuinely excelled across camera traps, drone footage, and specimen photographs simultaneously.

We therefore introduce \benchmarkname{}, a domain-grounded vision benchmark of \num{9} application-driven tasks that span \num{4} taxonomic kingdoms (animals, plants, fungi, and protozoa) captured from \num{6} distinct image distributions: drone footage, curated web video, microscope micrographs, in-situ RGB photos, RGB specimen images, and camera-trap frames (see \cref{sec:benchmark,tab:tasks} for more details).
The corpus contains \num{3.1}M images (\num{337} GB).
Evaluation reports macro-F1 for every task, with two tasks (FungiCLEF and FishNet) scored by their domain-standard metrics.
Each dataset downloads via a single-file Python script (fully documented).
Evaluation parallelizes across SLURM clusters or runs on a single GPU; ViT-B/16 and ViT-L/14 checkpoints finish in about one hour on an NVIDIA A6000, with larger models scaling predictably.

A linear fit over \nckpts{} pre-trained vision checkpoints underscores the need for new benchmarks.
Across all checkpoints, ImageNet explains only one-third of \benchmarkname{} variance ($R^{2}=0.34$) and agrees in rank just $\rho=0.55$, meaning the ImageNet-preferred model is actually worse on \benchmarkname{} roughly 22\% of the time.%
\footnote{$R^{2}$ has a \num{95}\% confidence interval of $[0.20,0.58]$; $\rho$ has a \num{95}\% confidence interval of $[0.45,0.64]$; both are significant with $p<0.0005$ via \num{5000}-perm randomization. Mis-ranking probability is $\frac{1}{2}(1-\rho)$.}
The mismatch widens at the frontier: among models above 75\% on ImageNet, rank concordance drops to $\rho=0.42$, so the supposed ``best'' model is mis-ranked \num{30}\% of the time.\footnote{$\rho$ for $>75\%$ has a \num{95}\% confidence interval of $[0.15,0.65]$ and is significant with $p<0.01$ via \num{5000}-perm randomization.}
These numbers make one conclusion unavoidable: web-photo leaderboards have ceased to be a trustworthy proxy for progress in scientific AI. 
\benchmarkname{} stands as a proof-of-concept, showing how domain workflows, long-tail metrics, and modality stress tests can be distilled into a single, open benchmark, and points the way toward equally realistic suites for medicine, manufacturing, and every other data-rich science.

\begin{figure}
    \centering
    \includegraphics[width=\linewidth]{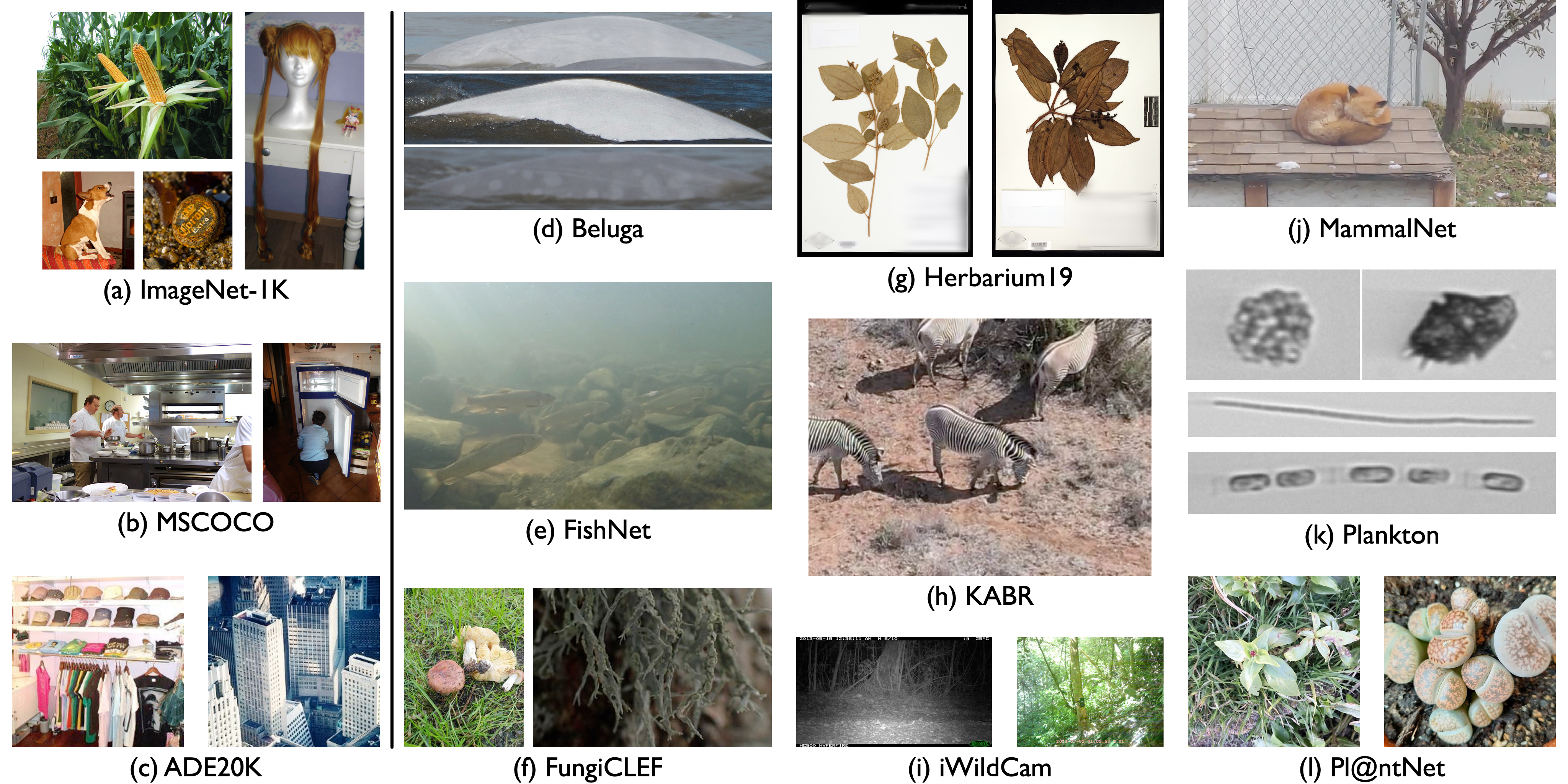}
    \caption{\textbf{Left (a-c):} Random example images from ImageNet-1K, MSCOCO and ADE20K, three popular general-domain vision benchmarks \citep{deng2009imagenet1k,lin2014mscoco,zhou2017ade20k}.
    \textbf{Right (d-l):} Random example images from each of the nine tasks in \benchmarkname{}.
    \textbf{Tasks in \benchmarkname{} have radically different image distributions compared to general-domain vision benchmarks.}
    }\label{fig:example-images}
\end{figure}

\section{Benchmark Suite \& Protocol}\label{sec:benchmark}
An effective ecological vision benchmark must address fundamental limitations in existing evaluation frameworks. 
First, it requires diversity across multiple dimensions: taxonomic breadth spanning microorganisms to mammals; varied image regimes from microscopy to camera traps; task diversity beyond simple classification; and natural class imbalances reflecting real-world species distributions. 
Second, it must balance proxy-driven tasks (measuring general capability) with mission-driven tasks (assessing operational utility for conservation applications). 
Third, it must provide rigorous statistical tools (confidence intervals, significance testing, and rank stability analysis) to distinguish genuine performance differences from benchmark lottery effects. 

Neither ImageNet-1K \citep{deng2009imagenet1k} nor iNat2021 \citep{van2021inat21newt} satisfies these requirements.
 
ImageNet lacks ecological diversity, while iNat2021 offers taxonomic breadth but limited task variety and no mission-driven evaluation. 
Most critically, our analysis reveals that once models exceed 75\% accuracy on ImageNet, the benchmark loses predictive power for ecological performance ($\rho$ drops from 0.82 to 0.55), rendering it insufficient as a proxy for ecological vision capability. \benchmarkname{} addresses these limitations through a minimal embedding interface that dramatically reduces integration overhead while providing comprehensive coverage across the ecological axes that matter most.

\textbf{Tasks.} \benchmarkname{} consolidates \num{9} public, application-driven tasks spanning \num{4} kingdoms (animals, plants, fungi, protists) and \num{6} image regimes (camera-trap RGB/IR, drone video frames, museum specimens, in-situ macro, web video, micrographs), totaling \num{3.1}M images. Tasks cover species ID, individual re-ID, behavior classification, and functional trait prediction. Example images are in \cref{fig:example-images} and task summaries are in \cref{tab:tasks}. 

\textbf{Implementation.} Models implement one contract $f:\text{image}\!\to\!\mathbb{R}^d$ (frozen embeddings). We fit linear or logistic probes per task, report macro-F1 by default (FishNet and FungiCLEF use task-specific metrics), and bootstrap confidence intervals. 

\textbf{Design Goals.} Embrace distributional diversity, evaluate long-tail class balance explicitly, and isolate representation quality from task-specific engineering via a uniform probing protocol.

\begin{table}[t]
    \centering
    \small
    \caption{Datasets across key dimensions that distinguish general computer vision benchmarks from ecological vision tasks. 
    $^*$\textit{Mission} tasks serve a specific ecological application (\cmark) rather than a general benchmark purpose (\xmark). 
    $^\dagger$\textit{Context} indicates whether images show organisms in their natural environment (in-situ) or as preserved specimens. ``Target'' indicates the classification target. 
    \textbf{Takeaway:} ImageNet-1K fundamentally differs from other ecological tasks because it is taxonomically unrestricted and web-scraped rather than scientifically curated. }
    \label{tab:tasks}
    \begin{tabular}{lccccc}
    \toprule
    Name & Mission?$^*$ & Taxon & Source & Context$^\dagger$ & Target \\
    \midrule
    ImageNet-1K & \xmark & - & Web-scraped & - & Object \\
    iNat2021 & \xmark & Diverse & Citizen science & In-situ & Species \\
    NeWT & \xmark & Diverse & Citizen science & In-situ & Varied \\
    \midrule
    BelugaID & \cmark & \textit{D. leucas} & Citizen science & In-situ & Individuals \\
    FishNet & \xmark & Fish & Natural collections & Specimen & Functional Traits \\
    FungiCLEF & \xmark & Fungi & Citizen science & In-situ & Species \\
    Herbarium19 & \cmark & Plants & Natural collections & Specimen & Species  \\
    iWildCam21 & \cmark & Mammals & Research studies & In-situ & Species \\
    KABR & \cmark & Mammals & Research study & In-situ & Behaviors \\
    MammalNet & \xmark & Mammals & Web-scraped & In-situ & Behaviors \\
    Plankton & \cmark & Protists & Research study & In-situ & Species \\
    Pl@ntNet & \xmark & Plants & Citizen science & In-situ & Species \\
    \bottomrule
    \end{tabular}
\end{table}

\section{Benchmark Results}

We evaluate \nckpts{} pre-trained vision models across \nfamilies{} model families on \benchmarkname{}.
We use a single Nvidia A6000 GPUs to evaluate all models; we will release both the individual model predictions and the aggregate statistics upon acceptance.
The results for each model family's top checkpoint are in \cref{tab:mini-results}; results for all checkpoints are available at \href{https://samuelstevens.me/biobench}{samstevens.me/biobench}.
Our analysis throughout this work considers all checkpoints.

Across \nckpts{} checkpoints, ImageNet-1K top-1 accuracy explains only $R^2=\num{0.34}$ of \benchmarkname{} variance; rank concordance is $\rho=\num{0.55}$ overall and drops further above \num{75}\% ImageNet (\cref{fig:hook}). 
Thus, the ImageNet-preferred model is worse on \benchmarkname{} \textbf{roughly $30\%$ of the time at the frontier}.

\begin{wrapfigure}{r}{0.4\textwidth}
    \centering
    \small
    \includegraphics[width=0.4\textwidth]{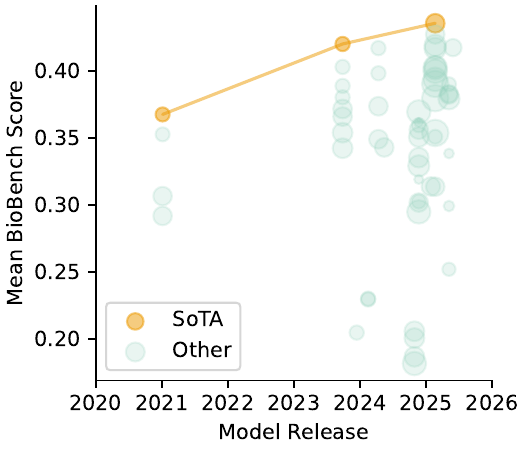}
    \vspace{-18pt}
    \caption{\benchmarkname{} scores over time. The majority of new models fail to improve on \benchmarkname{}.}\label{fig:timeline}
\end{wrapfigure}

We measure progress over \benchmarkname{} over time in \cref{fig:timeline}; despite general performance claims from many released generalist models, only CLIP \citep{radford2021clip}, SigLIP \citep{zhai2023siglip} and SigLIP 2 \citep{tschannen2025siglip2} set new state-of-the-art scores on \benchmarkname{}.

\begin{table}[t]
    \centering
    \small
    \setlength\tabcolsep{3pt}
    \caption{An overview of each model family's top-performing model on ImageNet-1K, NeWT and all tasks in \benchmarkname{}. State-of-the-art results for each task, along with their source, are reported at the bottom. ``-'' indicates no published state-of-the-art result. \textbf{Mean} is across all tasks in \benchmarkname{} (not ImageNet-1K or NeWT). 
    $^\dagger$Micro-accuracy (SOTA), not macro-F1 (ours).
    $^\ddagger$Macro-accuracy (SOTA), not macro-F1 (ours).
    }
    \label{tab:mini-results}
    \begin{tabular}{llr@{\hskip 12pt}rrrrrrrrrrrr}
        \toprule
        Family & Architecture & \verticalcell{Res. (px)} 
        & \verticalcell{ImageNet-1K} & \verticalcell{NeWT} 
        & \verticalcell{Beluga} & \verticalcell{FishNet} & \verticalcell{FungiCLEF} 
        & \verticalcell{Herbarium19} & \verticalcell{iWildCam} & \verticalcell{KABR} 
        & \verticalcell{MammalNet} & \verticalcell{Plankton} & \verticalcell{Pl@ntNet} & \verticalcell{\textbf{Mean}} \\
        
        \midrule
        CLIP & ViT-L/14 & 336 & 83.9 & 83.6 & 2.8 & 64.4 & 27.7 & 53.6 & 23.2 & 52.2 & 62.8 & 3.7 & 40.4 & 36.7 \\
        
        SigLIP & SO400M/14 & 384 & 87.8 & 86.0 & 4.0 & 69.0 & 38.6 & 63.7 & 25.7 & \textbf{59.3} & 66.3 & 4.0 & 47.4 & 42.0 \\
        SigLIP 2 & ViT‑1B/16 & 384 & \textbf{88.9} & 86.7 & 3.6 & 70.7 & 39.0 & 65.2 & 29.3 & \textbf{58.4} & \textbf{73.9} & 4.0 & 47.9 & \textbf{43.5} \\
        
        DINOv2 & ViT-g/14 & 224 & 86.7 & 82.8 & 4.5 & \textbf{75.2} & 34.2 & 64.3 & \textbf{30.5} & 53.7 & 57.1 & \textbf{4.2} & 51.5 & 41.7 \\
        
        AIMv2 & ViT-3B/14 & 448 & 86.7 & 84.0 & 1.7 & 59.2 & 34.4 & 48.3 & 20.5 & \textbf{58.9} & 68.8 & 4.0 & 36.7 & 36.9  \\
        
        SAM 2 & Hiera Large & 1024 & 33.9 & 64.2 & 3.1 & 45.8 & 16.1 & 12.7 & 5.4 & 38.5 & 33.7 & 3.2 & 9.5 & 18.7 \\
        
        V-JEPA & ViT-H/16 & 224 & 49.0 & 68.0 & \textbf{9.2} & 50.7 & 20.8 & 13.4 & 6.0 & 47.4 & 38.2 & 3.2 & 17.5 & 22.9 \\
        
        BioCLIP & ViT-B/16 & 224 & 58.5 & 82.7 & 4.6 & 62.6 & 40.6 & 52.6 & 17.2 & 46.1 & 35.7 & 3.8 & 45.4 & 34.3 \\
        BioCLIP 2 & ViT-L/14 & 224 & 80.0 & \textbf{89.1} & 3.0 & 71.8 & \textbf{51.0} & \textbf{73.1} & 24.7 & 48.0 & 46.4 & 3.9 & \textbf{53.8} & 41.7 \\
        BioTrove & ViT-B/16 & 224 & 45.3 & 82.9 & 3.7 & 59.7 & 41.6 & 47.0 & 11.1 & 37.3 & 30.0 & 3.8 & 48.1 & 31.4 \\
        MegaDesc. & Swin-L/4 & 384 & 49.9 & 71.3 & 8.0 & 50.2 & 22.1 & 14.0 & 6.9 & 32.3 & 31.1 & 2.1 & 17.6 & 20.5 \\
        \midrule
        \multicolumn{3}{l}{\textit{Random Prediction}} & 0.1 & 50.0 & 0.1 & 47.9 & 13.9 & 0.1 & 0.5 & 12.5 & 8.3 & 2.1 & 0.1 & 9.5 \\
        \multicolumn{3}{l}{\textit{Task-Specific State-of-the-Art}} & 91.0 & 80.6 & 66.5 & 81.7 & - & $^\dagger$89.9 & 66.7 & 65.8 & $^\ddagger$37.8 & - & - \\
        \multicolumn{3}{l}{\textit{[Source]}}
        & \citep{yu2022coca} & \citep{van2021inat21newt} & \citep{vcermak2024belugasota} & \citep{khan2023fishnet} & & \citep{tan2019herbarium19} & \citep{bahng2022iwildcamsota} & \citep{kholiavchenko2024deep} & \citep{chen2023mammalnet}  \\
        \bottomrule
    \end{tabular}
\end{table}

\section{Related Work}

General-domain benchmarks (ImageNet, COCO, ADE20K) catalyzed vision progress but are fragile under distribution shift and long-tail structure. Transfer suites such as VTAB and Taskonomy assess representation reuse but contain little ecological content.
Transfer suites such as VTAB \citep[spans 19 tasks across diverse domains][]{zhai2019vtab} or Taskonomy \citep[26 visual tasks][]{zamir2018taskonomy} assess representation reuse across domains. 
However, these benchmarks include minimal ecological content and fail to capture the specific challenges of biodiversity monitoring: fine-grained taxonomic distinctions, extreme environmental variability, and long-tailed species distributions.

iNaturalist \cite{van2021inat21newt} provides fine-grained species classification but doesn't incorporate temporal behavior or ecological trait prediction. 
Pl@ntNet \cite{garcin2021plantnet300k} focuses exclusively on plant identification. 
WILDS \cite{koh2021wilds} includes iWildCam \citep{beery2020iwildcam} for camera trap imagery but treats ecological monitoring as just one of many domains rather than exploring its multi-faceted challenges.

These isolated efforts highlight the critical need for \benchmarkname{}: conservation practitioners currently lack systematic guidance on which vision architectures best transfer to the complex, interconnected tasks comprising ecological monitoring workflows.

Methodological work \citep{gulrajani2021domainbed,dehghani2021benchmark} highlights the importance of consistent protocols—an ethos we adopt via a single embedding API, class-balanced metrics, and bootstrap uncertainty.

\section{Limitations \& Future Work}

While we argue that \benchmarkname{} meaningfully improves the state of ecological benchmarking and offers lessons applicable to other scientific domains, we have not explored every aspect.
\textbf{Limited Scope.} We focus on ecology; medicine and manufacturing may emphasize different tasks (e.g., detection/segmentation, calibration).
\textbf{Frozen features.} Probing isolates representation quality but underestimates task-specific fine-tuning gains.
\textbf{Metrics.} Macro-F1 rewards tail performance; some applications prefer operating-point metrics (e.g., precision@recall). 

\benchmarkname{} shows that ImageNet-driven model choice is unreliable for scientific imagery and offers a minimal, reproducible recipe to evaluate models where it matters. 
We hope \benchmarkname{} serves both as a practical guide for ecological workflows and as a template for building equally grounded benchmarks in other sciences.

\acksection{}

We thank Jianyang Gu, Tanya Berger-Wolf and Yu Su for their valuable feedback.
Our research is supported by NSF OAC 2118240.

\bibliography{main}

@inproceedings{deng2009imagenet1k,
  title={Imagenet: A large-scale hierarchical image database},
  author={Deng, Jia and Dong, Wei and Socher, Richard and Li, Li-Jia and Li, Kai and Fei-Fei, Li},
  booktitle={2009 IEEE conference on computer vision and pattern recognition},
  pages={248--255},
  year={2009},
  organization={Ieee}
}

@inproceedings{garcin2021plantnet300k,
    author={Garcin, Camille and Joly, Alexis and Bonnet, Pierre and Lombardo, Jean-Christophe and Affouard, Antoine and Chouet, Mathias and Servajean, Maximilien and Lorieul, Titouan and Salmon, Joseph},
    booktitle={NeurIPS Datasets and Benchmarks 2021},
    title={{Pl@ntNet-300K}: a plant image dataset with high label ambiguity and a long-tailed distribution},
    year={2021},
}

@inproceedings{khan2023fishnet,
    author    = {Khan, Faizan Farooq and Li, Xiang and Temple, Andrew J. and Elhoseiny, Mohamed},
    title     = {FishNet: A Large-scale Dataset and Benchmark for Fish Recognition, Detection, and Functional Trait Prediction},
    booktitle = {Proceedings of the IEEE/CVF International Conference on Computer Vision (ICCV)},
    month     = {October},
    year      = {2023},
    pages     = {20496-20506}
}

@article{tan2019herbarium19,
  title={The herbarium challenge 2019 dataset},
  author={Tan, Kiat Chuan and Liu, Yulong and Ambrose, Barbara and Tulig, Melissa and Belongie, Serge},
  journal={arXiv preprint arXiv:1906.05372},
  year={2019}
}

@inproceedings{kholiavchenko2024kabr,
  title={KABR: In-situ dataset for kenyan animal behavior recognition from drone videos},
  author={Kholiavchenko, Maksim and Kline, Jenna and Ramirez, Michelle and Stevens, Sam and Sheets, Alec and Babu, Reshma and Banerji, Namrata and Campolongo, Elizabeth and Thompson, Matthew and Van Tiel, Nina and others},
  booktitle={Proceedings of the IEEE/CVF Winter Conference on Applications of Computer Vision},
  pages={31--40},
  year={2024}
}

@article{kholiavchenko2024deep,
  title={Deep dive into KABR: a dataset for understanding ungulate behavior from in-situ drone video},
  author={Kholiavchenko, Maksim and Kline, Jenna and Kukushkin, Maksim and Brookes, Otto and Stevens, Sam and Duporge, Isla and Sheets, Alec and Babu, Reshma R and Banerji, Namrata and Campolongo, Elizabeth and others},
  journal={Multimedia Tools and Applications},
  pages={1--20},
  year={2024},
  publisher={Springer}
}

@article{bahng2022iwildcamsota,
  title={Exploring visual prompts for adapting large-scale models},
  author={Bahng, Hyojin and Jahanian, Ali and Sankaranarayanan, Swami and Isola, Phillip},
  journal={arXiv preprint arXiv:2203.17274},
  year={2022}
}

@inproceedings{vcermak2024belugasota,
  title={WildlifeDatasets: An open-source toolkit for animal re-identification},
  author={{\v{C}}erm{\'a}k, Vojt{\v{e}}ch and Picek, Lukas and Adam, Luk{\'a}{\v{s}} and Papafitsoros, Kostas},
  booktitle={Proceedings of the IEEE/CVF Winter Conference on Applications of Computer Vision},
  pages={5953--5963},
  year={2024}
}

@article{yu2022coca,
  title={Coca: Contrastive captioners are image-text foundation models},
  author={Yu, Jiahui and Wang, Zirui and Vasudevan, Vijay and Yeung, Legg and Seyedhosseini, Mojtaba and Wu, Yonghui},
  journal={arXiv preprint arXiv:2205.01917},
  year={2022}
}

@inproceedings{lin2014mscoco,
  title={Microsoft coco: Common objects in context},
  author={Lin, Tsung-Yi and Maire, Michael and Belongie, Serge and Hays, James and Perona, Pietro and Ramanan, Deva and Doll{\'a}r, Piotr and Zitnick, C Lawrence},
  booktitle={Computer vision--ECCV 2014: 13th European conference, zurich, Switzerland, September 6-12, 2014, proceedings, part v 13},
  pages={740--755},
  year={2014},
  organization={Springer}
}

@inproceedings{zhou2017ade20k,
  title={Scene parsing through ade20k dataset},
  author={Zhou, Bolei and Zhao, Hang and Puig, Xavier and Fidler, Sanja and Barriuso, Adela and Torralba, Antonio},
  booktitle={Proceedings of the IEEE conference on computer vision and pattern recognition},
  pages={633--641},
  year={2017}
}

@inproceedings{chen2023mammalnet,
  title={Mammalnet: A large-scale video benchmark for mammal recognition and behavior understanding},
  author={Chen, Jun and Hu, Ming and Coker, Darren J and Berumen, Michael L and Costelloe, Blair and Beery, Sara and Rohrbach, Anna and Elhoseiny, Mohamed},
  booktitle={Proceedings of the IEEE/CVF conference on computer vision and pattern recognition},
  pages={13052--13061},
  year={2023}
}

@article{weinstein2018computer,
  title={A computer vision for animal ecology},
  author={Weinstein, Ben G},
  journal={Journal of Animal Ecology},
  volume={87},
  number={3},
  pages={533--545},
  year={2018},
  publisher={Wiley Online Library}
}

@article{tuia2022perspectives,
  title={Perspectives in machine learning for wildlife conservation},
  author={Tuia, Devis and Kellenberger, Benjamin and Beery, Sara and Costelloe, Blair R and Zuffi, Silvia and Risse, Benjamin and Mathis, Alexander and Mathis, Mackenzie W and Van Langevelde, Frank and Burghardt, Tilo and others},
  journal={Nature communications},
  volume={13},
  number={1},
  pages={792},
  year={2022},
  publisher={Nature Publishing Group UK London}
}

@article{zech2018variable,
  title={Variable generalization performance of a deep learning model to detect pneumonia in chest radiographs: a cross-sectional study},
  author={Zech, John R and Badgeley, Marcus A and Liu, Manway and Costa, Anthony B and Titano, Joseph J and Oermann, Eric Karl},
  journal={PLoS medicine},
  volume={15},
  number={11},
  pages={e1002683},
  year={2018},
  publisher={Public Library of Science San Francisco, CA USA}
}

@article{raghu2019transfusion,
  title={Transfusion: Understanding transfer learning for medical imaging},
  author={Raghu, Maithra and Zhang, Chiyuan and Kleinberg, Jon and Bengio, Samy},
  journal={Advances in neural information processing systems},
  volume={32},
  year={2019}
}

@article{wightman2021resnet,
  title={Resnet strikes back: An improved training procedure in timm},
  author={Wightman, Ross and Touvron, Hugo and J{\'e}gou, Herv{\'e}},
  journal={arXiv preprint arXiv:2110.00476},
  year={2021}
}

@article{oquab2023dinov2,
  title={Dinov2: Learning robust visual features without supervision},
  author={Oquab, Maxime and Darcet, Timoth{\'e}e and Moutakanni, Th{\'e}o and Vo, Huy and Szafraniec, Marc and Khalidov, Vasil and Fernandez, Pierre and Haziza, Daniel and Massa, Francisco and El-Nouby, Alaaeldin and others},
  journal={arXiv preprint arXiv:2304.07193},
  year={2023}
}

@inproceedings{radford2021clip,
  title={Learning transferable visual models from natural language supervision},
  author={Radford, Alec and Kim, Jong Wook and Hallacy, Chris and Ramesh, Aditya and Goh, Gabriel and Agarwal, Sandhini and Sastry, Girish and Askell, Amanda and Mishkin, Pamela and Clark, Jack and others},
  booktitle={International conference on machine learning},
  pages={8748--8763},
  year={2021},
  organization={PmLR}
}

@inproceedings{zhai2023siglip,
  title={Sigmoid loss for language image pre-training},
  author={Zhai, Xiaohua and Mustafa, Basil and Kolesnikov, Alexander and Beyer, Lucas},
  booktitle={Proceedings of the IEEE/CVF international conference on computer vision},
  pages={11975--11986},
  year={2023}
}

@article{fini2024aimv2,
  title={Multimodal autoregressive pre-training of large vision encoders},
  author={Fini, Enrico and Shukor, Mustafa and Li, Xiujun and Dufter, Philipp and Klein, Michal and Haldimann, David and Aitharaju, Sai and da Costa, Victor Guilherme Turrisi and B{\'e}thune, Louis and Gan, Zhe and others},
  journal={arXiv preprint arXiv:2411.14402},
  year={2024}
}

@article{fang2023does,
  title={Does progress on ImageNet transfer to real-world datasets?},
  author={Fang, Alex and Kornblith, Simon and Schmidt, Ludwig},
  journal={Advances in Neural Information Processing Systems},
  volume={36},
  pages={25050--25080},
  year={2023}
}

@inproceedings{vishniakov2024beyondimagenet,
  title = {ConvNet vs Transformer, Supervised vs CLIP: Beyond ImageNet Accuracy},
  author = {Vishniakov, Kirill and Shen, Zhiqiang and Liu, Zhuang},
  year = 2024,
  booktitle = {Proceedings of the 41st International Conference on Machine Learning (ICML)},
  url = {https://icml.cc/virtual/2024/poster/34818},
}

@inproceedings{koh2021wilds,
  title={Wilds: A benchmark of in-the-wild distribution shifts},
  author={Koh, Pang Wei and Sagawa, Shiori and Marklund, Henrik and Xie, Sang Michael and Zhang, Marvin and Balsubramani, Akshay and Hu, Weihua and Yasunaga, Michihiro and Phillips, Richard Lanas and Gao, Irena and others},
  booktitle={International conference on machine learning},
  pages={5637--5664},
  year={2021},
  organization={PMLR}
}

@article{beery2020iwildcam,
    title={The iWildCam 2020 Competition Dataset},
    author={Beery, Sara and Cole, Elijah and Gjoka, Arvi},
    journal={arXiv preprint arXiv:2004.10340},
    year={2020}
}

@article{algasov2024beluga,
  title={Understanding the Impact of Training Set Size on Animal Re-identification},
  author={Algasov, Aleksandr and Nepovinnykh, Ekaterina and Eerola, Tuomas and K{\"a}lvi{\"a}inen, Heikki and Stewart, Charles V and Otarashvili, Lasha and Holmberg, Jason A},
  journal={arXiv preprint arXiv:2405.15976},
  year={2024}
}

@inproceedings{vcermak2024wildlifedatasets,
  title={WildlifeDatasets: An open-source toolkit for animal re-identification},
  author={{\v{C}}erm{\'a}k, Vojt{\v{e}}ch and Picek, Lukas and Adam, Luk{\'a}{\v{s}} and Papafitsoros, Kostas},
  booktitle={Proceedings of the IEEE/CVF Winter Conference on Applications of Computer Vision},
  pages={5953--5963},
  year={2024}
}

@inproceedings{rolnick2024applicationdriven,
  title = {Position: Application-Driven Innovation in Machine Learning},
  author = {Rolnick, David and Aspuru-Guzik, Alan and Beery, Sara and Dilkina, Bistra and Donti, Priya L. and Ghassemi, Marzyeh and Kerner, Hannah and Monteleoni, Claire and Rolf, Esther and Tambe, Milind and White, Adam},
  year = 2024,
  month = {21--27 Jul},
  booktitle = {Proceedings of the 41st International Conference on Machine Learning},
  publisher = {PMLR},
  series = {Proceedings of Machine Learning Research},
  volume = 235,
  pages = {42707--42718},
  url = {https://proceedings.mlr.press/v235/rolnick24a.html},
  editor = {Salakhutdinov, Ruslan and Kolter, Zico and Heller, Katherine and Weller, Adrian and Oliver, Nuria and Scarlett, Jonathan and Berkenkamp, Felix}
}

@inproceedings{van2021inat21newt,
  title={Benchmarking representation learning for natural world image collections},
  author={Van Horn, Grant and Cole, Elijah and Beery, Sara and Wilber, Kimberly and Belongie, Serge and Mac Aodha, Oisin},
  booktitle={Proceedings of the IEEE/CVF conference on computer vision and pattern recognition},
  pages={12884--12893},
  year={2021}
}

@article{dosovitskiy2020vit,
  title={An image is worth 16x16 words: Transformers for image recognition at scale},
  author={Dosovitskiy, Alexey and Beyer, Lucas and Kolesnikov, Alexander and Weissenborn, Dirk and Zhai, Xiaohua and Unterthiner, Thomas and Dehghani, Mostafa and Minderer, Matthias and Heigold, Georg and Gelly, Sylvain and others},
  journal={arXiv preprint arXiv:2010.11929},
  year={2020}
}

@article{tschannen2025siglip2,
  title={Siglip 2: Multilingual vision-language encoders with improved semantic understanding, localization, and dense features},
  author={Tschannen, Michael and Gritsenko, Alexey and Wang, Xiao and Naeem, Muhammad Ferjad and Alabdulmohsin, Ibrahim and Parthasarathy, Nikhil and Evans, Talfan and Beyer, Lucas and Xia, Ye and Mustafa, Basil and others},
  journal={arXiv preprint arXiv:2502.14786},
  year={2025}
}

@article{zhai2019vtab,
  title={The visual task adaptation benchmark},
  author={Zhai, Xiaohua and Puigcerver, Joan and Kolesnikov, Alexander and Ruyssen, Pierre and Riquelme, Carlos and Lucic, Mario and Djolonga, Josip and Pinto, Andre Susano and Neumann, Maxim and Dosovitskiy, Alexey and others},
  year={2019}
}

@inproceedings{zamir2018taskonomy,
  title={Taskonomy: Disentangling task transfer learning},
  author={Zamir, Amir R and Sax, Alexander and Shen, William and Guibas, Leonidas J and Malik, Jitendra and Savarese, Silvio},
  booktitle={Proceedings of the IEEE conference on computer vision and pattern recognition},
  pages={3712--3722},
  year={2018}
}

@article{dehghani2021benchmark,
  title={The benchmark lottery},
  author={Dehghani, Mostafa and Tay, Yi and Gritsenko, Alexey A and Zhao, Zhe and Houlsby, Neil and Diaz, Fernando and Metzler, Donald and Vinyals, Oriol},
  journal={arXiv preprint arXiv:2107.07002},
  year={2021}
}

@inproceedings{gulrajani2021domainbed,
title={In Search of Lost Domain Generalization},
author={Ishaan Gulrajani and David Lopez-Paz},
booktitle={International Conference on Learning Representations},
year={2021},
url={https://openreview.net/forum?id=lQdXeXDoWtI}
}

\end{document}